# Accelerating Surgical Robotics Research: A Review of 10 Years with the da Vinci Research Kit

Claudia D'Ettorre[1*], Andrea Mariani[2*], Agostino Stilli[1], Ferdinando Rodriguez y Baena[3], Pietro Valdastri[4], Anton Deguet[5], Peter Kazanzides[5], Russell H. Taylor[5], Gregory S. Fischer[6], Simon P. DiMaio[7], Arianna Menciassi[2] and Danail Stoyanov[1]

*Abstract*— **Robotic-assisted surgery is now well-established in clinical practice and has become the gold standard clinical treatment option for several clinical indications. The field of robotic-assisted surgery is expected to grow substantially in the next decade with a range of new robotic devices emerging to address unmet clinical needs across different specialties. A vibrant surgical robotics research community is pivotal for conceptualizing such new systems as well as for developing and training the engineers and scientists to translate them into practice. The da Vinci Research Kit (dVRK), an academic and industry collaborative effort to re-purpose decommissioned da Vinci surgical systems (Intuitive Surgical Inc, CA, USA) as a research platform for surgical robotics research, has been a key initiative for addressing a barrier to entry for new research groups in surgical robotics. In this paper, we present an extensive review of the publications that have been facilitated by the dVRK over the past decade. We classify research efforts into different categories and outline some of the major challenges and needs for the robotics community to maintain this initiative and build upon it.**

## I. INTRODUCTION

Robotics is at the heart of modern healthcare engineering. Robotic-assisted surgery in particular has been one of the most significant technological additions to surgical capabilities over the past two decades [1]. With the introduction of laparoscopic or minimally invasive surgery (MIS) as an alternative to traditional open surgery, the decoupling of the surgeon's direct access to the internal anatomy generates the need to improve ergonomics and creates favorable arrangement for robotic tele-manipulator support. In MIS, the visceral anatomy is accessed through small trocar made ports using specialized elongated instruments and a camera (i.e., laparoscope) to observe the surgical site. Robotic-assisted MIS (RMIS) uses the same principle but the tools and the scope are actuated by motors and control systems providing enhanced instrument dexterity and precision, as well as immersive visualization at the surgical console. The most successful and widely used RMIS platform, the da Vinci surgical system (Intuitive Surgical Inc. (ISI), Sunnyvale, CA, USA), is shown in Fig.1 (left). To date, more than 5K da Vinci surgical system have been deployed worldwide performing over 7M surgical procedures across different anatomical regions [2]. Urology, gynecology and general surgery represent the main application areas where the da Vinci surgical system has been used although many other specializations have also developed robotic approaches, for example in thoracic and transoral surgery [3] (Fig. 1, right).

The impact on both clinical science and engineering research of the da Vinci surgical system has also been significant, with more than 25K peer-reviewed articles reported, as shown in Fig. 1 (right). Many clinical studies and case reports belong to this body of literature and focus on investigating the efficacy of RMIS or its development for new approaches or specialties. In addition to clinical research, the da Vinci surgical system has also facilitated many engineering publications and stimulated innovation in surgical robotics technology. In the early years since the clinical introduction of the robot, such engineering research was predominantly focused on the development of algorithms that utilize data from the system, either video or kinematic information, or external sensors adjunct to the main robotic platform. However, relatively few institutions had da Vinci surgical systems available for research use, the majority of platforms were dedicated to clinical utilization, and kinematic information was accessible through an API which required a research collaboration agreement with ISI. This inevitably restricted the number of academic or industry researchers able to contribute to advancing the field.

To address the challenges in booting surgical robotics research, the da Vinci Research Kit (dVRK) research platform was developed through a collaboration between academic institutions, Johns Hopkins University and Worcester Polytechnic Institute, and ISI in 2012 [4]. Seminal papers

[*] These authors equally contributed to this work.
[1] Claudia D'Ettorre, Agostino Stilli and Danail Stoyanov are with the Wellcome/EPSRC Centre for Interventional and Surgical Sciences (WEISS), University College London, London W1W 7EJ, UK (e-mail: c.dettorre@ucl.ac.uk).
[2] Andrea Mariani and Arianna Menciassi are with the BioRobotics Institute and the Department of Excellence in Robotics & AI of Scuola Superiore Sant'Anna (SSSA), Pisa, Italy.
[3] Ferdinando Rodriguez y Baena is with Mechanical Engineering Department, Imperial College London, UK.
[4] Pietro Valdastri is with the STORM Lab, Institute of Robotics, Autonomous Systems and Sensing, School of Electronic and Electrical Engineering, University of Leeds, Leeds, UK.
[5] Anton Deguet, Peter Kazanzides and Russell H. Taylor are with Johns Hopkins University, Baltimore, USA.
[6] Gregory S. Fischer is with the Automation and Interventional Medicine Laboratory, Worcester Polytechnic Institute, Worcester, MA, USA.
[7] Simon DiMaio is with Intuitive Surgical Inc, Sunnyvale, California, USA.

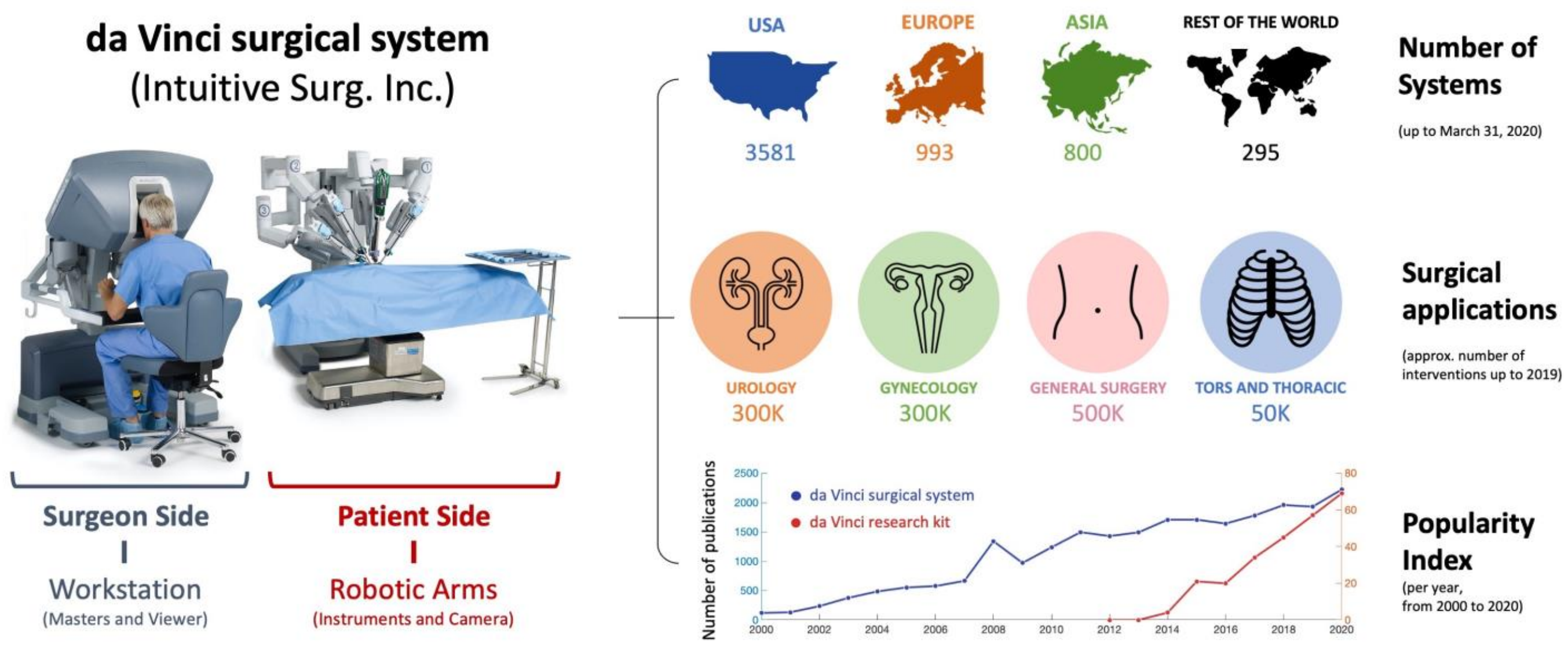

Fig. 1. (left) The da Vinci surgical system is a surgical tele-manipulator: the surgeon sits at a workstation and controls instruments inside the patient by handling a couple of masters; (right top) global distribution of da Vinci surgical systems in 2020; (right middle) surgical specialties and total number of interventions up to 2019 using the da Vinci surgical system; (right bottom, blue curve) number of publications citing the da Vinci surgical system as found in Dimensions.ai [11] looking for the string “da Vinci Surgical System” in the Medical, Health Sciences and Engineering fields; (right bottom, red curve) number of publications citing the da Vinci Research Kit (dVRK) as found in Dimensions.ai [11] looking for the string “da Vinci Research Kit”.

[5],[6] where the platform was presented for the first time, outline the dVRK and its mission. The idea behind the dVRK initiative is to provide the core hardware, i.e., a first-generation da Vinci surgical system, to a network of researchers worldwide, by repurposing retired clinical systems. This hardware is provided in combination with dedicated electronics to create a system that enables researchers to access to any level of the control system of the robot as well as the data streams within it. The dVRK components are the master console (the interface at the surgeon side), the robotic arms to handle the tools and the scope at the patient side, and the controller boxes containing the electronics (Fig. 2). To date, the dVRK, together with the purely research focused RAVEN robot [7] are the only examples of open research platforms in surgical robotics that have been used across multiple research groups. The introduction of the dVRK allowed research centers to share a common hardware platform without restricted access to the underlying back- and forward control system. This has led to a significant boost to the development of research in surgical robotics during the last decade and generated new opportunities for collaboration and to connect a surgical robot to other technologies. Fig. 1 (bottom, right) shows the increasing number of publications citing and using the dVRK.

With this paper, we aim to provide a comprehensive overview of the research carried out to date using the dVRK. We hope to help readers to quickly understand the current activities of the community and the possibilities enabled by the open access architecture. It is our view that the impact of the system should be a precedent for similar initiatives between industry-academic consortia.

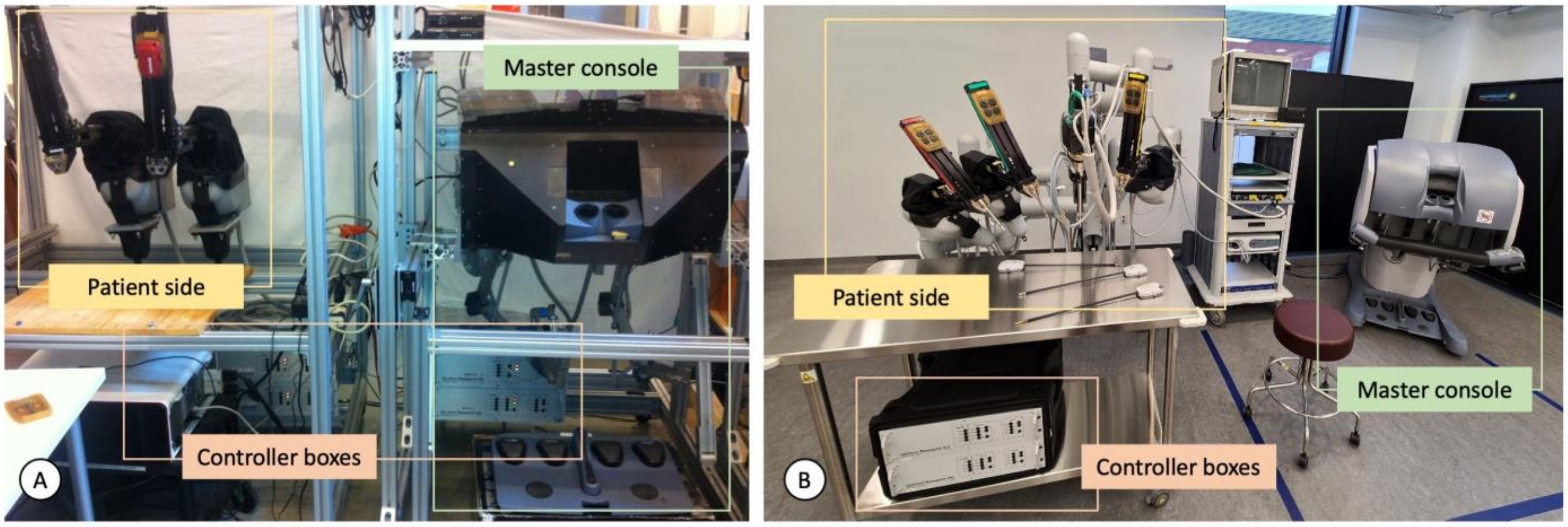

Fig. 2. The da Vinci Research Kit (dVRK) is available as the collection and integration of spare parts from the first-generation da Vinci surgical system (subfigure A, on the left, from Johns Hopkins University) or as the full retired first-generation da Vinci surgical system (subfigure B, on the right, from Worcester Polytechnic Institute). All the dVRK platforms feature the same main components: the patient side, i.e., the robotic arms to handle the surgical tools; the master console, i.e., the interface at the surgeon side; the controller boxes containing the electronics that guarantee accessibility and control of the system. The former version (subfigure A, on the left) does not include the endoscopic camera and its robotic manipulator at the patient side.

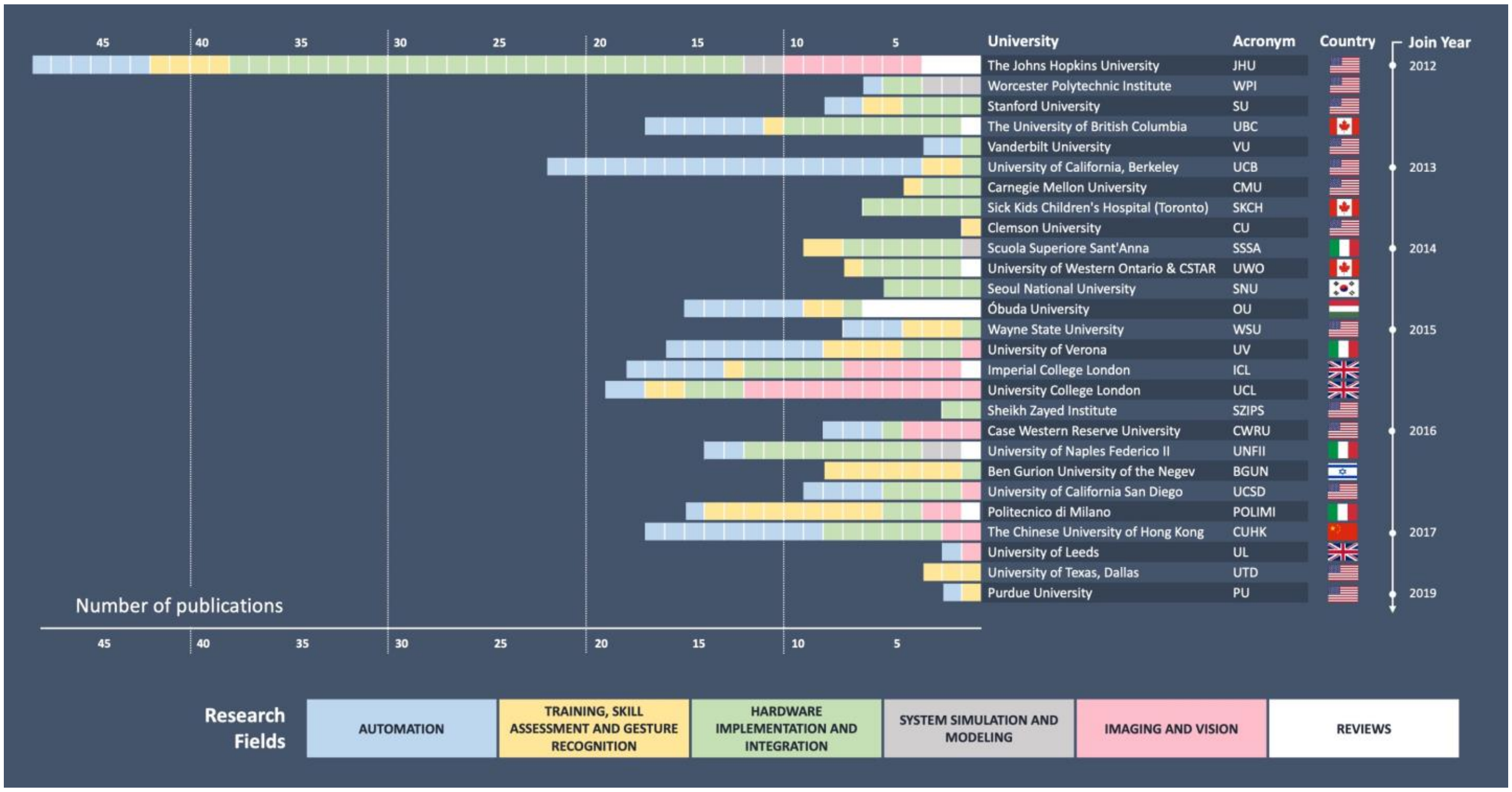

Fig. 3. This histogram shows the publications associated to the dVRK community members. All the research centers are listed in temporal order based on their joining year. They feature name, acronym and respective country. The left side of the graph represents the number of publications for each research center. Each square represents a single publication. The color code is used to classify the topic of the paper corresponding to each square according to its research field, whose legend is reported on the bottom.

## II. SEARCH PROTOCOL

The dVRK community is currently composed of 40 research centers from more than 10 different countries. The initiative is US led, starting in 2012 with the later addition of research sites in Europe and Asia. The full timeline and list of research centers can be found at [4], [9]. Today, the dVRK consortium includes mostly universities and academic centers within hospitals, and some companies (i.e., Surgnova [8] and of course ISI who support and underpin the entire initiative with their technology [9]).

Our review focuses only on scientific publications rather than research resulted in patents. In order to identify and catalog all the available publications involving the dVRK, we followed a protocol querying three main databases: the dVRK Wiki Page [4], Google Scholar [10] and Dimensions.ai [11]. The PRISMA flow diagram associated to our search and selection can be found in the Appendix section (Fig. 6). All the papers published in international conferences or journals were taken into account, as well as all the publications related to workshops or symposiums, and the open-access articles stored in arXiv [12].

Firstly, we manually visited the research centers' websites as listed on the dVRK Wiki [4]. Whenever the link was active, papers were collected from the lab's website; if inactive, the name of the principal investigator was used to locate the laboratory website and the relative available list of publications. This first refined research generated a cluster of 142 publications.

We then extended this collection with the results from Google Scholar [10] with the query "da Vinci Research Kit". The research time interval was set between 2012 (origin of the dVRK community [4]) and 2020 producing 523 results. The results were further processed and refined by removing outliers where the dVRK was not actually mentioned in the *Methods* section of the work (that means it was just cited but not used in the experimental work), as well as filtering out master theses, duplicates and the works where the full text of the paper in English was not available online. This research finally generated 266 papers.

The last paper harvesting search was performed on Dimensions.ai [11] looking for the same "da Vinci Research Kit" string, generating 394 results. The same paper filtering, as carried out for the results from Google Scholar, was performed resulting in 270 publications. At this stage, these three screened datasets of papers (i.e., from the dVRK Wiki, Google Scholar and Dimensions.ai) have been cross-checked in order to ensure no duplications in the final collection of dVRK-related papers. 296 publications were obtained as final number.

In Fig. 3, the dVRK community members (for which at least one publication was found) are shown. They are listed on a timeline indicating the year they received the dVRK system following the same order of [4]. In case of publications involving multiple centers, the publication was assigned to the principal investigator's affiliation. In case of collaborations between dVRK community members and institutes external to the community, the publication was assigned to the dVRK community member.

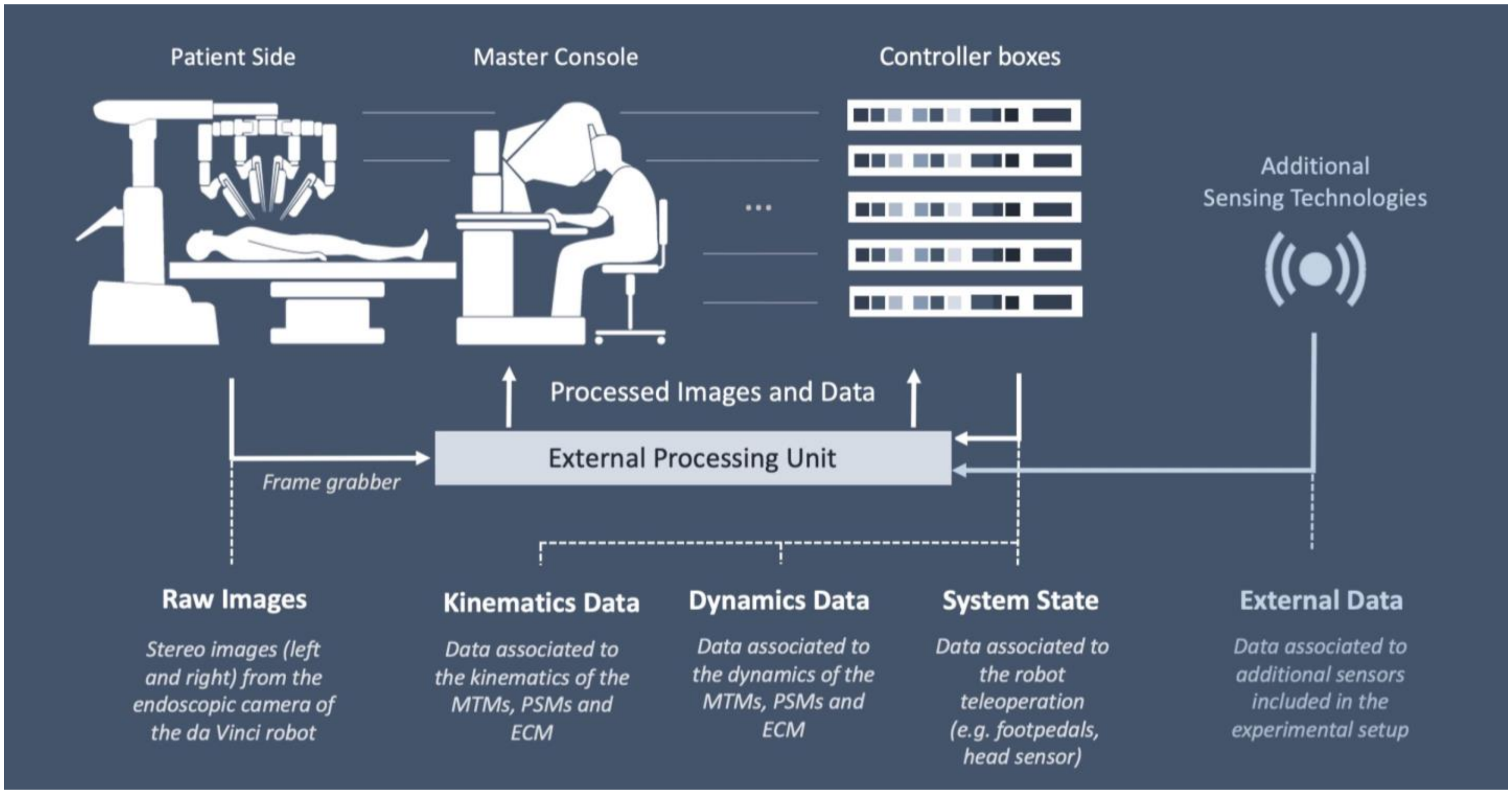

Fig. 4. Top – Sketch of the da Vinci Research Kit components. From left to right: patient side with the three patients side manipulators (PSM) and endoscopic camera manipulator (ECM); the master console including the foot pedal tray, the two master tool manipulators (MTM) and two high resolution stereo-viewers; the controller boxes and the vision elements (camera control units, light source). Bottom – Description of data types. These types of data that can be read (arrows entering the *External Process Unit*) and written (arrows exiting the *External Process Unit*) using the dVRK.

## III. PAPER CLASSIFICATION - RESEARCH FIELDS AND DATA TYPES

For analyzing the body of publications, six research fields were used for clustering: *Automation*; *Training, skill assessment and gesture recognition*; *Hardware implementation and integration*; *System simulation and modelling*; *Imaging and vision*; *Reviews*. These broadly categorize the published works though notably some works may involve multiple fields or be at the interface between fields. In the histogram of Fig. 3, each colored box corresponds to a publication of the related research field. A second clustering criteria to classify publications relies on five different data types, shown in Fig. 4 (bottom). The classes were defined based on the data used and/or collected to underpin the papers. The five different data types are: *Raw Images (RI),* i.e. the left and right frames coming from the da Vinci stereo endoscope or any other cameras. *Kinematics Data (KD)* and *Dynamics Data (DD)*, i.e. all the information associated to the kinematics and dynamics of the console side of the dVRK - Master Tool Manipulators (MTMs), as well as the instrument side - Patient Side Manipulators (PSMs) and Endoscopic Camera Manipulator (ECM). *System Data (SD),* i.e. the data associated to the robot teleoperation states, as signals coming from foot pedals, head sensor for operator presence detection, etc. *External data (ED),* a category that groups all the data associated with additional sensors that were connected and integrated with the dVRK platform in experimental test rigs, such as eye trackers, different imaging devices and sensors. Because of the importance of data and its utilization, especially with artificial intelligence (AI), this second categorization adds an important perspective to the work underpinned through the dVRK.

Table I reports the proposed classification highlighting both clustering categorizations.

### A. *Automation*

There is a large spectrum of opportunity for automating aspects of RMIS [308]: some of them may be already existing features such as tremor reduction; others are more forward-looking, such as the automation of an entire surgical task, where a clinician must rely on the robot for the execution of the action itself.

Automation in RMIS is always a combination of multiple areas of robotics research: robot design and control, medical image/sensing and real-time signal processing, and AI and machine learning. This category of dVRK research includes 81 publications, representing one of the most popular research areas that has benefitted from a system where algorithms can be used on hardware. There are different approaches that can be used to automate surgical tasks, for example involving a human in a preplanning stage, utilizing control theory to follow a human during the operation, or use machine learning techniques examples and execute them autonomously later.

We decided to group efforts in RMIS automation based on the aim of the proposed control strategy, as general control, instrument control and camera control.

*General control:* several efforts focus on developing new high-level control architectures for automation in RMIS without specializing on task-oriented applications [148], [169], [174], [291]. From focusing their attention to human-robot interaction approaches [65], [168], to general motion compen-TABLE I - Classification of the dVRK publications: on the horizontal axis, the five research macro areas are listed. Each area is then subdivided into five subgroups according to the type of the data used in the publication (RI – Raw Images, KD – Kinematics Data, DD – Dynamics Data, SD – System Data, ED – External Data). The sixth column is dedicated to the publications reviewing dVRK-related technologies.

| | Automation | | | | | Training, Skill Assessment and Gesture Recognition | | | | | Hardware Implementation and Integration | | | | | System Simulation and Modelling | | | | | Imaging and Vision | | | | | Reviews |
|---|---|---|---|---|---|---|---|---|---|---|---|---|---|---|---|---|---|---|---|---|---|---|---|---|---|---|
| | RI | KD | DD | SD | ED | RI | KD | DD | SD | ED | RI | KD | DD | SD | ED | RI | KD | DD | SD | ED | RI | KD | DD | SD | ED | |
| JHU | [14][15][16][17][18] | [14][15][16][17][18][19] | [14][15][16] | [14][15][16][18][19] | [14][15][16][18][19] | [22][23] | [20][22][23] | [21] | [22] | [20][21][22] | [5][6][24][25][26][27][28][29][30][31][32][33][34][35][36][37][38][42][43][44][45][46] | [5][6][13][24][25][26][27][28][29][30][31][32][33][34][35][36][37][38][39] [42][43] [44][46] | [5][6][13][24][27][28][29][30][31][32][33][34][35][36] [37][38][39][40][42][43] [44][46] | [5][6][13][24][27][28][29][30][31][32][33][34][35][36] [37][38][39][40][41] [42][43] [45][46] | [24][25][26][28][30][30][31][32][34] [37][38][40][42] [43][45][46] | [47][48] | [47][48] | [48] | [47] | [47] | [49][50][51][52][53][54][55] | [49][50][51][52][53][54][55] | [51][52][53][54][55] | [51][52][54][55] | [49][50][51][53][54][55] | [56][57][58] |
| WPI | | [59] | [59] | | | | | | | | [60] | [60][61] | [60][61] | [60][61] | | | [62] | [62][63][64] | | | | | | | | |
| SU | [65][66] | | [65] | [66] | | [67] | [67][68] | [67][68] | [68] | [67][68] | | [69][70][71][72] | [70][71][72] | [69] | [69][70][71][72] | | | | | | | | | | | |
| UBC | [73][74][75][77][78] | [73][75][76][77][78] | [75] | [75][76] | [73][75][76][77][78] | [79] | [79] | | [79] | [79] | [80][81][82][83][84][85][86][87] | [80][81][82][83][84][85][86][87][88] | [88] | [87] | [84][85][86][87][88] | | | | | | | | | | | [89] |
| VU | [90][91] | [90][91] | | | | | | | | | [92] | [92] | [92] | | [92] | | | | | | | | | | | |
| UCB | [93][94][95][96][97][98][99][100][101][102][103][104][105][106][107] [108][110][111] | [93][94][95][96][97][98][99][100][101][102][103][104][105][106][107] [108][109] [110][111] | [94][101][103] | [101] | [96][97][98][99][100][101][111] | [112][113] | [112][113] | | | | | [114] | [114] | | [114] | | | | | | | | | | | |
| CMU | | | | | | [115] | [115] | | | | [116][117] | [116] [117] | | | [118] | | | | | | | | | | | |
| SKCH | | | | | | | | | | | [119] | [119][120][121][122][123][124] | [119] [122][123][124] | | [119] [121][124] | | | | | | | | | | | |
| CU | | | | | | | | | | [125] | | | | | | | | | | | | | | | | |
| SSSA | | | | | | [126] | [126] | [126] | [126] | | | [127][128][129][130][131][132] | [127][128][130][131] | [127][129][132] | [127][129][130][131][132] | | [133] | [133] | | | | | | | | |
| UWO | | | | | | | [134] | [134] | | [134] | [135][136][137] | [135][136][137][138][139] | [137][138][139] | | [135][138] | | | | | | | | | | | [140] |
| SNU | | | | | | | | | | | [141][142][143][144][145] | [141][142][143][144][145] | [143][144] | [141][142][143][144] | [142][143][144][145] | | | | | | | | | | | |
| OU | [146][147][148][149][150][151] | [146][147][148][149][150][151] | | [148][149][151] | [148][151] | | | | | [152][153] | [154] | [154] | [154] | | [154] | | | | | | | | | | | [155][156][157][158][159][160] |
| WSU | [161][162][163] | [161][162][163] | | [163] | | [164][165] | [164][165] | [165] | [164][165][166] | [166] | [167] | [167] | [167] | | [167] | | | | | | | | | | | |
| UV | [168][169][170][171][172][173][174][175] | [168][169][170][171][172] [173][174][175] | [168][169][170][171][173][174] | [173] | [168][169][172][173][175] | [176][177][178][179] | [176][177][178][179] | | | [176][177][179] | [180] [181][182] | [180] [181][184] [182] | | [180] [181][184] | [181][182] | | | | | | [183] | [183] | | [183] | [183] | |
| ICL | [185][186][187][188][189] | [185][186][187][188][189] | | [186][187][189] | [185][186][188][189] | [190] | [190] | | | | [191][192][193][196] | [193][194][195] [196] | [192][194][195] | [191][192][193] | [191][193][194][195][196] | | | | | | [197][198][199][200][201] | [198][199][200][202] | | [199][200] | [199][200][202] | [203] |
| UCL | [204][205] | [204][205] | | | | [206][207] | [206][207] | | | | [208][209][210] | [208][209][210] | | [208][209] | [208][209][210] | | | | | | [211][212][213][214][215][216][217][218][219][220][221][222] | [213][214][215][216][221] | | [221] | | |
| SZIPS | | | | | | | | | | | | [223][224] | [223][224] | | | | | | | | | | | | | |
| CWRU | [225][226][227] | [225][226][227] | | | [225][226][227] | | | | | | | [228] | [228] | | [228] | | | | | | [229][230][231][232] | [229] [232] | | | [232] | |
| UNFII | [233] | [233][234] | [233][234] | | [233][234] | | | | | | [235][236][237][238][239][240][244] | [235][236][238][239][240][241][242][243][244] | [235][239][240][241][242][243][244] | | [240][242][243] [244] | | [245][246] | [245][246] | [245] | | | | | | | [247] |
| BGUN | | | | | | [248][249][253][254] | [248][249][250][251][252][253][254] | [252] | [252][253] | [253][254] | | [255] | [255] | | [255] | | | | | | | | | | | |
| UCSD | [256][257][258][259] | [256][257][258] [259] | [258] | [258] | | | | | | | [261][262][263] | [261][262][263][264] | [263] | | [264] | | | | | | [265] | [265] | | | [265] | |
| POLIMI | [266] | [266] | | [266] | | [267][268][272][274] | [267][268][269][270][271][272][273][274][275][276] | [267][270][272][273][275] | [267][268][269][270][271][272][273][274][275][276] | | [277] | [277][278] | [279] | [279] | [277] | | | | | | [280][281] | [280][281] | | [280][281] | | [282] |
| CUHK | [283][284][285][286][287][288][290][291] | [283][284][285][286][287][288][289] [290][291] | [289] | [283][284][285][288][289] | [283][289][290][291] | | | | | | [292][293][298] | [292][293][295][296][297] [298] | [295][296][297] | | [292][293][294] [298] | | | | | | [299][300] | [299] | | | [299] | |
| UL | [301] | [301] | | [301] | | | | | | | | | | | | | | | | | [302] | | | | | |
| UTD | | | | | | [303][304] | [303][304][305] | | | [305] | | | | | | | | | | | | | | | | |
| PU | [306] | [306] | [306] | [306] | | | | | | [307] | | | | | | | | | | | | | | | | |

sation [73], or control considering uncertainties [234].

*Instrument control*: this section groups all the contributions that have been made towards the attempt of automation of specific surgical subtasks. Six main tasks appear to be targets widely investigated for automation. For the *suturing task*, including works related to knot tying and needle insertion, we reported the following: [77], [78], [93], [96], [97], [185], [225], [226], [227], [233], [285], [17], [288], [289]. The *pick, transfer and place task* was mainly characterized by experiments relying on pegs and rings from the Fundamentals of Laparoscopic Surgery (FLS) training paradigm [310] ([59], [75], [76], [99], [104], [111],[170], [187], [171], [74], [172], [306]) or new surgical tools [204]. A lot of the remaining works focus on tissue interaction. This application category includes papers working on *cutting and debridement* [66], [95], [98], [100], [103], [105], [110], [259]. As well as *retraction and dissection* of tissues [109], [146], [147], [149], [151], [175], [189], [301] or blood suction [257], [258]. Also *tissue palpation* for locating tumors or vessels and more general *tissue manipulation* as in [14], [15], [16], [18], [90], [91], [94], [106], [107], [256], [283], [188], [173], sometimes just using common fabric [102], [108].

*Camera control*: additional literature included studies that investigated how to control the endoscopic camera or assist in controlling it. In RMIS, the surgeon can switch between controlling the tools and the camera through a pedal clutch interface. This acts as a safety mechanism to ensure that joint motion, which can be risky, is prevented but the transition typically leads to a segmented workflow, where the surgeon repositions the camera in order to optimize the view of the workspace. Investigations on how to optimize the camera control in order to minimize the time lost in repositioning the camera have been a longstanding effort focused on autonomous navigation of the endoscope [19], [101], [142], [150], [266], [284].

### B. Training, skill assessment and gesture recognition

This research field encompasses all the publications focusing on gesture learning and recognition utilizing different data sources to infer surgical process, for a total of 46 publications. Surgical robots, like all the surgical instrumentation, require extensive, dedicated training to learn how to operate precisely and safely. Robotics with the additional encoder information compared to normal instrumentation (specifically an open platform such as the dVRK) open attractive opportunities to study motor learning: as haptic interfaces, robots provide easy access to the data associated to the operator's hand motion. This information (mainly kinematics and dynamics) can be used to study gestures, assess skills, and improve learning by training augmentation.

*Training platforms and augmentation*: several studies propose the development of training platforms (in dry lab [20], [22], and simulation [21], [272], as well as training protocols (based on data from expert surgeons [79], [134], [165], or introducing autonomous strategies that can adapt the training session to the trainee [276], [269], [271], [305]). Among these protocols, haptic guidance and virtual fixtures (i.e. the application of forces to the trainee's manipulators to guide and teach the correct movement) have been of particular interest [251], [252], [270], [275].

*Skill assessment:* as a fundamental component of training, skill assessment has received attention (focusing on proficiency analysis [23], [115], [166], [176], [179], [253], [254], [267], [303], [304], as well as addressing the mental and physical workload of the user [153], [307], and the influence of training on haptic perception [67]).

*Workflow analysis*: gesture analysis [68], [126], [152], [164], [250] and segmentation [112], [113], [125], [164], [177], [178], [191], [207], [208], [273], have been also widely investigated in the research community, both for image segmentation and augmentation.

### C. Hardware implementation and integration

Hardware implementation and integration is the most heterogenous category, hence the highest number of publications (111) belong to this group.

*dVRK platform implementation and integration:* this group includes all the works published during the development of the dVRK. Both the hardware and software components are described in [5], [6], [13], [24], [27], [35], [36], [39], [40], [41], [42], [43]. Few new integrations were lately published in [297].

*Haptics and pseudo-haptics*: several research groups have investigated how to overcome the lack of haptic feedback in the current da Vinci system. Numerous hardware and software applications [70], [71], [72], [87], [88], [118], [119], [127], [139], [223], with eventual links to automation [131], [242], [243], [255], [262], are the main contributions to the force sensing integration with the dVRK. Related to this topic, the use of virtual fixtures, previously mentioned in Section B, as an intra-operative guiding tool, has been investigated in [28], [29], [30], [85], [92], [192], [193], [235], [238], [239], [278]. Furthermore, research works focused on augmented reality to provide the surgeons with visual feedback about forces (the so called pseudo-haptics) have been presented in [31], [32], [34], [116], [117], [261], [277], [298].

*New surgical tools*: another group of publications includes all those works focusing on the design and integration of new tools compatible with the dVRK: new surgical instruments [120], [121], [122], [123], [124], [128], [129], [127], [144], [145], [184], [224], [236], [237], [244] ,[293], [294], [296], and new sensing systems [81], [82], [208], [228], [292].

*New control interfaces*: few works focused on the development of novel control interfaces of the endoscopic camera [44], [141], [142], [143], [167] and new flexible endoscopes and vision devices [286], [287], as well as novel master interfaces [191], [194], [195].

*Surgical workflow optimization*: the last large subgroup of publications in this research area is related to the implementation and integration with the dVRK of technologies that can enhance the surgeon's workflow and perception, such

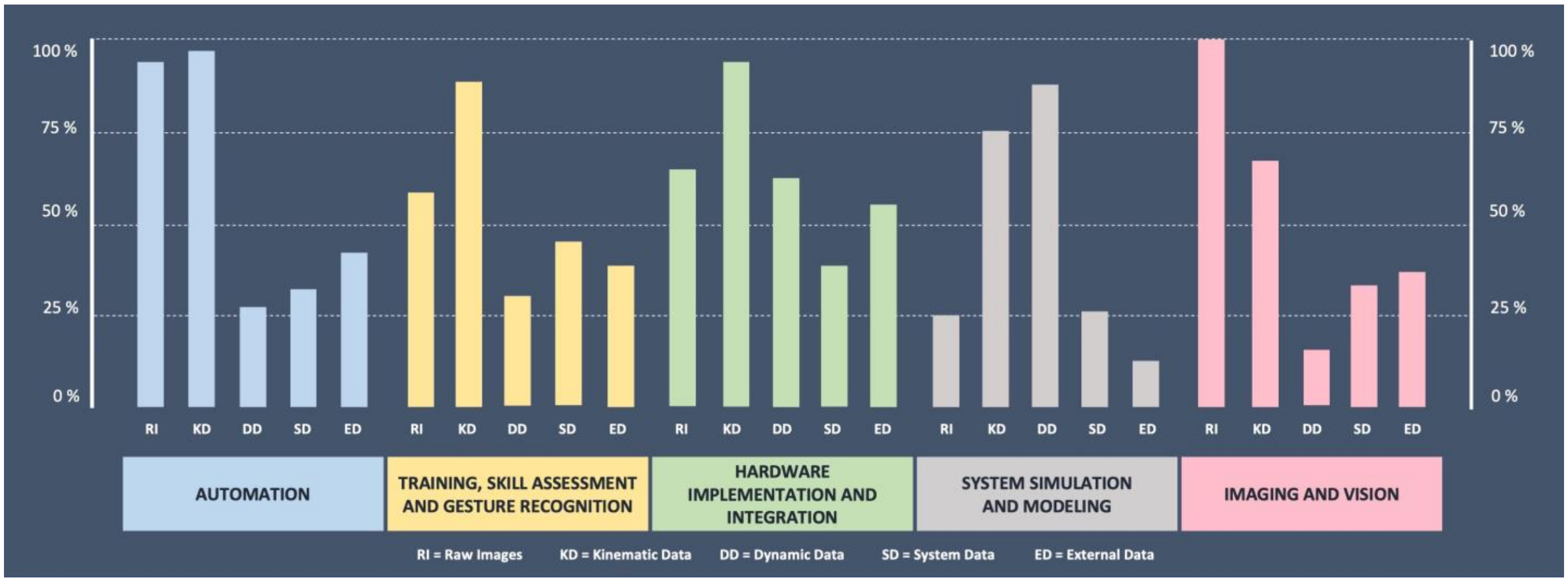

Fig. 5. Histogram of data usage (in percentage) for each category based on the publications coming from TABLE I. The percentage refers to the number of publications involving a certain data type out of the total number of publications in a certain research field.

as [33], [80], [84], [86], [114], [135], [136], [137], [182], [184], [196], [209], [210], [241], [264], [279], [295]. A significant research effort has been done also for improving the teleoperation paradigm such as in [25], [37], [38], [45], [46], [60], [69].

*Other:* additional works investigate the use of the dVRK as basis platform to explore clinical indications or non-clinical applications beyond the current intent for the clinical da Vinci system. For example in retinal surgery [26], heart surgery [83], portable simulators [240], and using the master controllers to drive vehicles in simulations [154].

### D. System simulation and modelling

This smaller group of 8 publications contains all the studies that focused on the integration of the dVRK into simulation environments to obtain realistic robot interaction with rigid and soft objects [47], [63], [245]. In this framework, the identification of the kinematics and dynamics properties of the robotic arms have been addressed [48], [62], [64], [133], [246]. The size of this research field is limited since all the works using simulation environments as tools to implement other solutions (e.g., for testing task automation, or as a training environment) have been classified in the specific category of application.

### E. Imaging and Vision

This category includes 36 publications related to the processing of the images and video coming from the dVRK's stereo laparoscopic camera. A wide range of vision algorithms are applied and developed to this data with publications ranging from investigations of the detection of features in the images (to perform camera calibration, tissue and instrument tracking or image segmentation) to systems enabling overlays of additional information onto the images displayed by the scope for augmented reality.

*Camera calibration*: this first group includes publications investigating approaches for endoscope to surgical tools registration (i.e. hand-eye calibration) [50], [183],[199], [200], [213], [214], [299], [232], as well as determining the camera intrinsic parameters using dVRK information [49].

*Segmentation*: works aimed at detecting, segmenting and tracking important elements in the surgical scene, such as surgical instruments [198], [211], [212], [215], [217], [218], [220], [221], [222], [229], [249], [265],[300], suturing needles [231], tissues [302] and suturing threads [230]. The dVRK has been important in this area for developing open datasets especially for instrument detection and segmentation as well as pose estimation.

*Augmentation*: other works rely on different or emerging imaging modalities and techniques like ultrasound or photoacoustic imaging [51], [55] to implement image guidance [54] to enhance surgical capabilities and patient safety during operations [53], [216], [280], [281]. In [52] the segmentation of a marker is used as control of a 4-DOF laparoscopic instrument. In [201], [219], images are used to learn how to estimate the depth and 3D shape of the workspace, or how to automatically remove smoke from the surgeons' field of view [197].

### F. Reviews

Several major review publications cite the dVRK and study the literature in RMIS related topics. Comprehensive reviews on the state of the art of RMIS and future research directions have been presented in [57], [58], [140], [156], [158], [159], [160], [282]. Works like [155], [157] review the general aspects of autonomy in robotic surgery, while [89] and [247] focus on human aspects in control and robotic interaction. In [203] the legal implications of using AI for automation in surgical practice are discussed, while virtual and augmented reality in robotic surgery are reviewed in [56].

## IV. DISCUSSION

This review paper focuses on describing the state of the art as we approach the first decade of the dVRK by providing a comprehensive collection of the papers that have been published so far in a wide range of research topics. Overall, 296 papers have been classified into five different categories. In each category, each publication was then classified also based on the type of data it relied on because one of the main advantaged the dVRK has offered is the use of a surgical robot for data generation. As a summary, Fig. 5 shows the percentage usage of a given type of data for each research field.

Starting from the automation research category, almost all the papers we reviewed rely on the use of endoscopic images and/or KD from the encoders. This trend obviously persists in the imaging and vision classes with research outputs based on KD being slightly less active. For training and skill assessment and surgical gesture recognition most papers rely on KD, using any other type of data in less than 50% of the cases or exploiting ED. When it comes to hardware implementation and integration almost all the types of data are greater than 50%, preserving a good balance except for the KD. For system simulation and integration, it is possible to notice how KD and DD are used in the vast majority of publications, leaving the other data type to less than 25%. In general, the correlation between the type of data and each application area shows the increasingly importance of images in RMIS, since in almost all the categories RI crosses the 50%. The extensive use of KD and DD also highlights the importance of having a research platform, as the dVRK, that facilitates the ability to exploit the robot as a haptic interface and to make use of the systems' data generation capabilities. Furthermore, the open-access design of the dVRK incentivizes and enables researchers to integrate it with different types of hardware and software, as shown by the extensive usage of external data in almost all of the classes.

By taking consideration of the data usage in the published research and the research categories for the dVRK we hope to map the worldwide research activity the system has stimulated. Despite the non-exhaustive nature of this review report and analysis, we believe that the information collected provides a compelling account of the research areas and directions explored and enabled through the dVRK. It offers adopters of the dVRK a comprehensive overview of the research outputs, synopsis of activity of the different consortium stakeholders across the globe.

The review has highlighted the importance and impact that dVRK data generation and availability has had on stimulating research. This is not surprising because of the huge surge in activity in data intensive research in machine learning, computer vision and artificial intelligence in general. A future improvement for the dVRK platform would be enabling researchers to collect and store synchronized system data with minimum effort so that it can be used for different applications as well as providing basis for benchmarks and challenges. For example, all the experiments carried out in papers around surgeon training and skill assessment could be recorded in centralized data storage and used as demonstration to train algorithms for task automation. This also links to clinical data availability and areas of active development with research institutions under research agreements with ISI where data can be recorded from the clinical setting (using custom recording tools such as the dVLogger by ISI, like in [312]).

Multiple new initiatives can build on and evolve the dVRK's current capabilities and also can spawn additional development. An interesting addition considering the recent thrust in the automation area would be to invest significant effort and develop and integrate a fully-fledged simulation environment for research. This would open opportunities for researchers with the possibility to develop and test algorithms that require a vast number of learning iterations in reinforcement learning or unsupervised learning strategies. Additionally, a simulation dVRK would allow research teams without the space or hardware support infrastructure to work in the field. A connection between such a simulator and real systems with community development of the libraries and facilities for teleoperation could also be a exiting capability to explore further.

In summary of our review of the impact the dVRK has had on robotics research, we note a strong trend towards more effective data utilization in surgical robotic research is related to the possibility of making research platforms more compliant and open to the integration of different systems, in order to facilitate data collection, storage, sharing and usage. The work facilitated by the dVRK highlights this current area of development. However, the dVRK also does much more, with examples of significant effort and development facilitated by the platform in new hardware, integration with imaging or other non-robotic capabilities, and human factors studies. It is the authors' opinion that the platform has been a huge catalyst to research acceleration in RMIS and hopefully to the transition of research efforts into clinically meaningful solutions in future years. Maintaining the spirit of the dVRK both in terms of underpinning system and community will have continued impact on surgical robotics research. It will be extremely valuable to continue the initiative and see future generations of the da Vinci system become part of the research ecosystem the dVRK has created as their clinical use becomes retired or decommissioned.

## APPENDIX

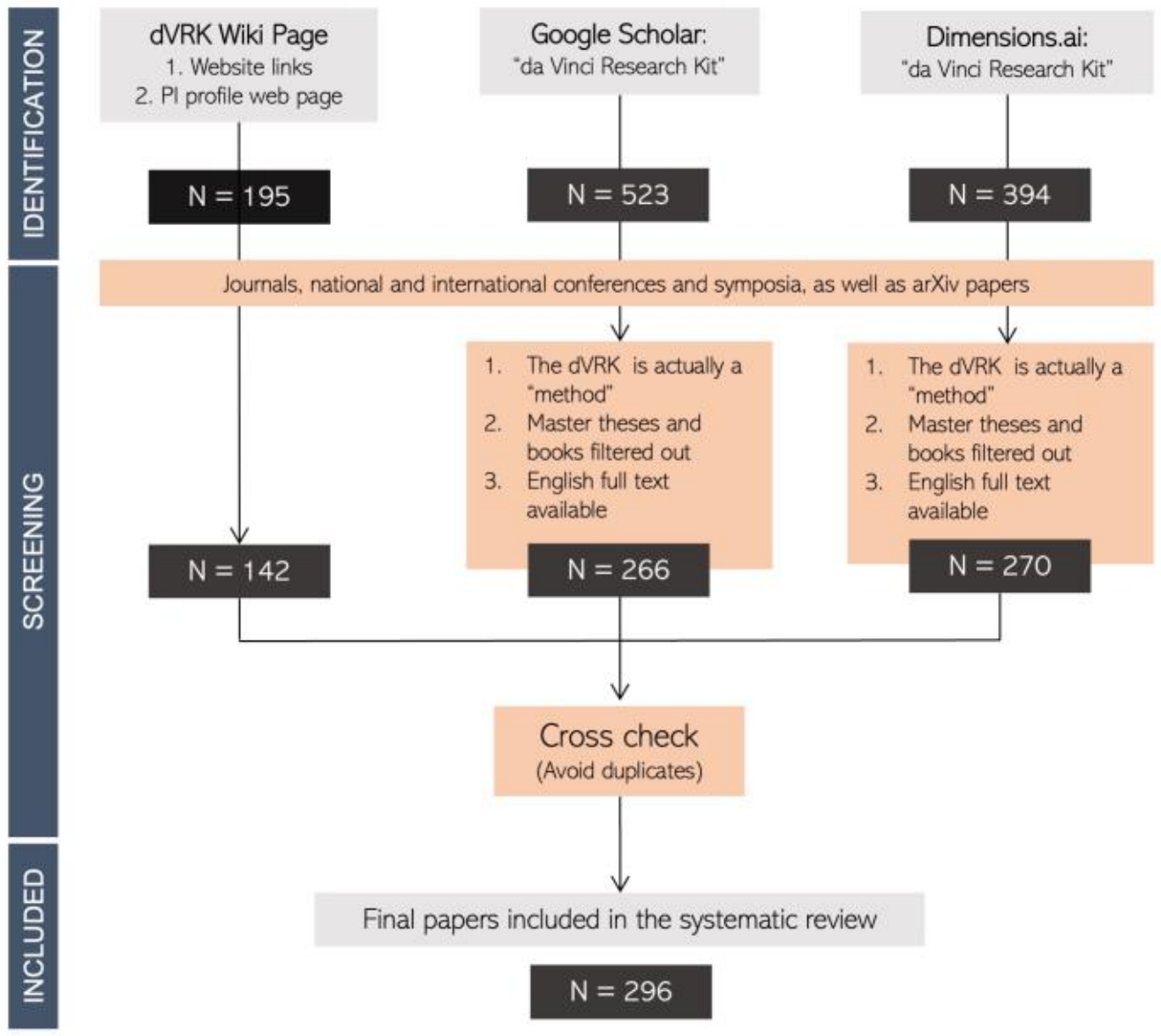

Fig. 6 - PRISMA flow diagram associated to the paper search and selection of this systematic review.